\documentclass[11pt,a4paper]{article}

\usepackage[margin=2.25cm]{geometry}
\usepackage{fontspec}
\usepackage{amsmath,amssymb}
\usepackage{booktabs,tabularx,array}
\usepackage{graphicx}
\usepackage[labelfont=bf,font=small]{caption}
\usepackage{float}
\usepackage{microtype}
\usepackage{xcolor}
\usepackage[colorlinks=true,allcolors=blue!55!black]{hyperref}

\graphicspath{{figures/}}
\newcommand{\tsdf}{\textsc{TSDF}}

\begin{document}

\begin{center}
{\LARGE\bfseries THA-Flow Generative Model:\\[3pt]
Prosthesis Geometry Prediction from Preoperative CT\par}
\vspace{0.8em}
{\large Yiping Wang\textsuperscript{1,*}, Jie Li\textsuperscript{2},
Jingyu Shen\textsuperscript{1}, Liao Wang\textsuperscript{2,*}\par}
\vspace{0.3em}
{\small \textsuperscript{1}Changzhou Jinse Medical Information Technology Co., Ltd.,
Changzhou 213000, Jiangsu, China\par}
{\small \textsuperscript{2}Shanghai Key Laboratory of Orthopaedic Implants,
Department of Orthopaedic Surgery, Shanghai Ninth People's Hospital,
Shanghai Jiao Tong University School of Medicine, Shanghai, China\par}
\vspace{0.3em}
{\small \textsuperscript{*}Correspondence: Yiping Wang
(\href{mailto:yiping.wang@mdivi.cn}{yiping.wang@mdivi.cn}) or Liao Wang
(\href{mailto:wang821127@163.com}{wang821127@163.com})\par}
\end{center}

\section*{Abstract}
Preoperative planning for total hip arthroplasty (THA) is commonly framed as selecting a single prosthesis configuration and placement for a patient's osseous anatomy. In practice, however, the same anatomy may admit several clinically reasonable solutions, making planning inherently a one-to-many problem that is better represented by a conditional probability distribution. We present THA-Flow, a conditional flow-matching model that generates three-dimensional prosthesis geometry directly from preoperative CT. Separate AutoencoderKL models compress preoperative bone anatomy and prosthesis geometry, while a three-dimensional UNet learns a rectified flow from Gaussian noise to the prosthesis latent space under spatial bone conditioning and optional structured prosthesis parameters. The retrospective cohort comprised 1,355 hips from 1,149 patients undergoing primary THA. Following rigid registration of postoperative CT to preoperative CT, the actual postoperative prostheses were transformed independently according to the pelvic and femoral registrations and represented as a dual-channel truncated signed distance field. The prosthesis autoencoder achieved a peak signal-to-noise ratio of 47.11 dB and a structural similarity index of 0.9964 on the validation set. Complete acetabular and femoral geometries were generated across seven major stem models representing 93.4\% of the cohort. Repeated bone-conditioned sampling preserved component position, alignment, and the principal bone--prosthesis interfaces while allowing limited local geometric variation. To our knowledge, THA-Flow represents the first application of generative AI to three-dimensional surgical planning for THA.

{\small\noindent\textbf{Keywords:} Total hip arthroplasty; Surgical planning; Patient-matched prosthesis; Flow-matching model\par}

\section{Introduction}
Preoperative planning for THA requires selection of prosthesis shape and size together with its spatial placement relative to patient anatomy. Current digital workflows rely primarily on two- or three-dimensional measurements, heuristic rules, and matching against commercial prosthesis libraries\cite{colombi2019planning}. CT-based systems such as AIHIP apply deep learning to pelvic and femoral segmentation and anatomical landmark detection, but component selection and placement remain downstream procedures based on measurements, explicit planning rules, and predefined prosthesis inventories\cite{chen2022aihip}. These systems improve the efficiency of image processing and measurement, yet still converge on a single component configuration and placement and therefore do not represent the distribution of plausible plans for the same anatomy.

Generative models offer a different formulation by permitting repeated sampling from a conditional distribution. THA-Net established the feasibility of synthesizing simulated postoperative radiographs from preoperative radiographs, with both unconditional and prosthesis-specified generation\cite{rouzrokh2024thanet}. Its output, however, remains a two-dimensional synthetic image. Generating the entire postoperative image also redraws bone, soft tissue, and the prosthesis interface, so realistic image appearance does not guarantee preservation of the patient's original osseous anatomy. Conditioning explicitly on the preoperative bone while restricting the target to three-dimensional prosthesis geometry allows the prediction to be superimposed directly on the unaltered patient anatomy.

The primary obstacle is the construction of spatially consistent three-dimensional supervision. Hip posture differs between preoperative and postoperative CT, while femoral neck resection and metal artefacts further undermine direct registration. The postoperative prosthesis can provide voxel-level supervision only after its acetabular and femoral components have been returned independently to the preoperative coordinate space with the pelvis and femur. High-resolution three-dimensional volumes also impose substantial memory and computational demands; latent-space generation reduces this burden while preserving the principal spatial structure\cite{rombach2022ldm,wang2025meddiffusion}. MAISI and related work have further demonstrated spatially controlled generation of large three-dimensional medical images, with rectified flow reducing the number of sampling iterations\cite{guo2025maisi,zhao2026maisiv2}. Compared with synthesizing a complete postoperative CT containing metal artefacts, soft-tissue texture, and complex acquisition-specific features, generating only an implicit representation of the target prosthesis geometry provides a more clearly defined and tractable objective. The output can also be converted directly into a surface for subsequent measurement, retrieval, and geometric analysis.

This study focuses on the three-dimensional generative formulation and technical validation of THA-Flow. Its contributions are threefold: a postoperative-to-preoperative registration pipeline that excludes metal-contaminated and surgically altered regions to establish reliable supervision in preoperative space; a dual-channel \tsdf{} representation in which acetabular and femoral components follow the pelvis and femur, respectively; and a latent rectified-flow model in which the bone latent is injected as a spatial condition from the first convolutional layer, with optional stem model and size as additional semantic controls. We evaluate generated geometry across stem designs, at the bone--prosthesis interface, and through conditional distribution consistency; latent reconstruction and velocity-field errors serve as technical checks on model training.

\section{Results}
\subsection{Conditional Inference Architecture}
\begin{figure}[H]
\centering
\includegraphics[width=\textwidth]{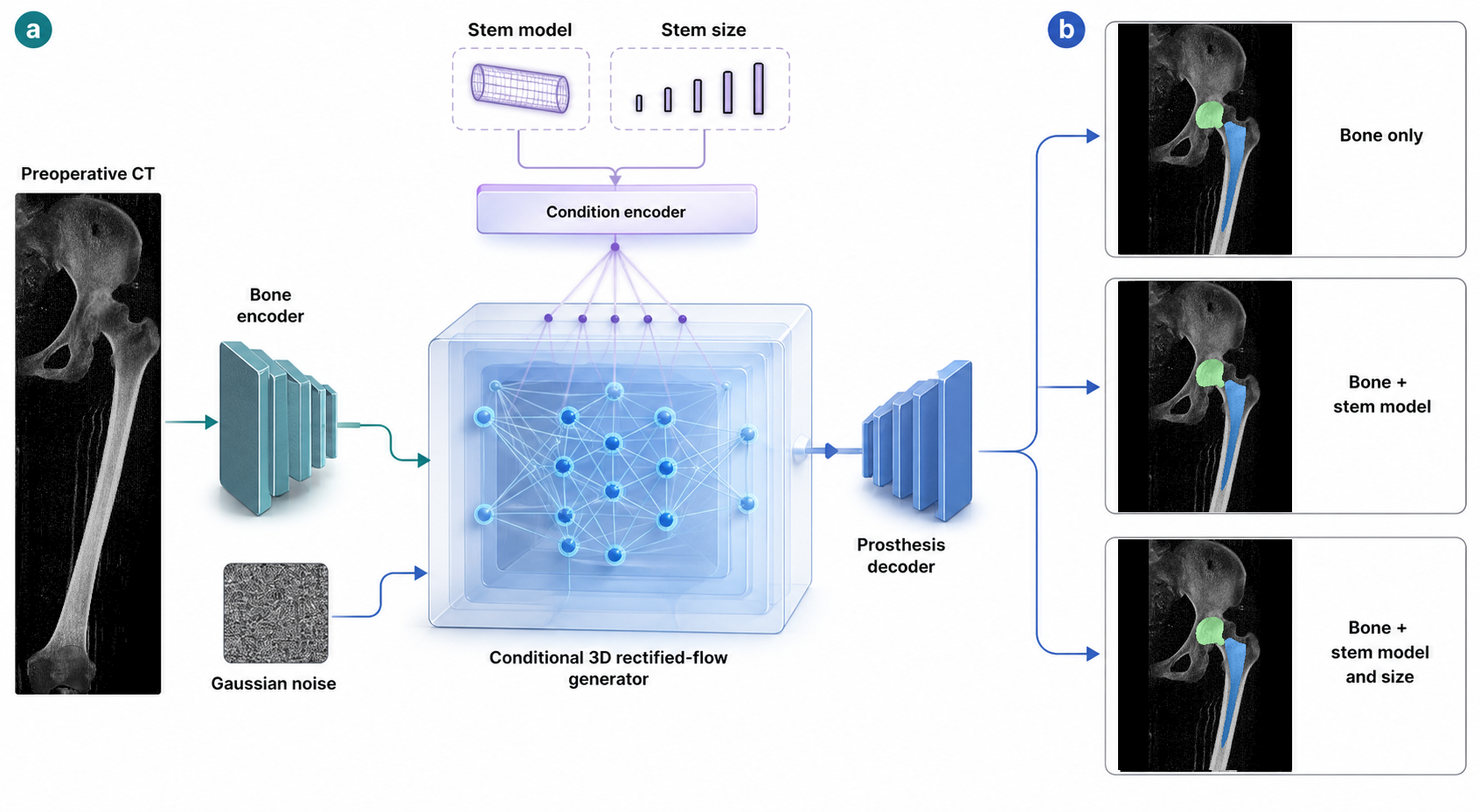}
\caption{\textbf{Conditional inference architecture and generation modes of THA-Flow.} \textbf{a}, Preoperative CT is encoded as a spatial bone condition, while the optional femoral stem model and size are encoded as semantic constraints. The three-dimensional generator transports a random latent state to a prosthesis latent, which is decoded in preoperative CT space. \textbf{b}, Outputs for the same patient under bone-only, bone-plus-model, and bone-plus-model-and-size conditions. Green denotes the acetabular cup and femoral head; blue denotes the femoral stem, separated from the head at the neck.}
\label{fig:inference-architecture}
\end{figure}

In the intended planning workflow, THA-Flow takes the preoperative CT as its primary input (Fig.~\ref{fig:inference-architecture}). A bone encoder first extracts anatomy-preserving spatial features, while random noise initializes the prosthesis state. When a femoral stem model or size is specified, a condition encoder converts these inputs into additional semantic constraints. The three-dimensional generator integrates the anatomy, optional product information, and evolving prosthesis state, and the prosthesis decoder then reconstructs three-dimensional geometry in the coordinate system of the preoperative CT for direct inspection against the patient's bone.

The same model supports three clinically relevant modes. With preoperative CT alone, it produces a patient-matched candidate from the bone geometry. Adding the stem model constrains the generated shape toward the specified product family. Adding both model and size further restricts stem dimensions while preserving anatomical fit. Figure~\ref{fig:inference-architecture} shows all three outputs for the same patient. Component position, alignment, and overall extent remain anchored by the bone condition, whereas the product condition modulates local stem geometry. On the single-GPU system used in this study, median latent integration time across the three modes was 0.31 s, and median generation time including sliding-window decoding was 1.76 s per case, supporting interactive candidate generation; model loading and preprocessing were excluded.

The discontinuity at the neck in Fig.~\ref{fig:inference-architecture} is a deliberate consequence of the supervisory coordinate system rather than a generation failure. The actual postoperative prosthesis is registered to preoperative CT with the pelvis and femur independently. The model therefore learns the geometry associated with each bone in the pathological preoperative joint configuration, not the assembled construct after joint reduction. Keeping the cup and femoral head in one channel preserves the postoperative hip centre and head--cup spatial relationship, avoiding separate estimation of their narrow clearance and relative pose. Separating this channel from the stem at the neck decouples patient-matched prosthesis generation from joint reduction and assembly. The two channels are nevertheless generated jointly in a shared coordinate system and can therefore encode their assembly relationship implicitly. Their relative displacement at the neck also provides a direct reference for post-processing: its longitudinal component represents the leg-length correction required to move from the preoperative pathological posture to the reduced construct.

\subsection{Patient-Matched Generation Across Stem Designs}
The seven most prevalent femoral stem models accounted for 1,265 hips, or 93.4\% of the cohort. One validation case from each model was used to compare the preoperative CT, the actual postoperative prosthesis registered to preoperative space, bone-only generation, and generation conditioned jointly on bone and the recorded stem model (Fig.~\ref{fig:cross-model}). Both generative modes produced complete acetabular and femoral geometry, with stem position and alignment adapting to the proximal femur and medullary canal. Model conditioning additionally constrained stem length, proximal width, and lateral profile without disrupting anatomical fit, demonstrating that a single generator can accommodate multiple widely used product families.

A bone-only result need not reproduce the stem design selected at the actual operation, yet it can retain a plausible intramedullary position, alignment, and spatial extent and thus represent another candidate permitted by the anatomy. Conditioning on the recorded stem model shifts the generated geometry toward the corresponding design features. The postoperative prosthesis remains one clinically implemented solution rather than the sole correct answer for preoperative planning. Pairwise agreement with that solution is informative about model behaviour, but should not be the only criterion for the clinical plausibility of a generated plan.

\begin{figure}[H]
\centering
\includegraphics[width=0.80\textwidth]{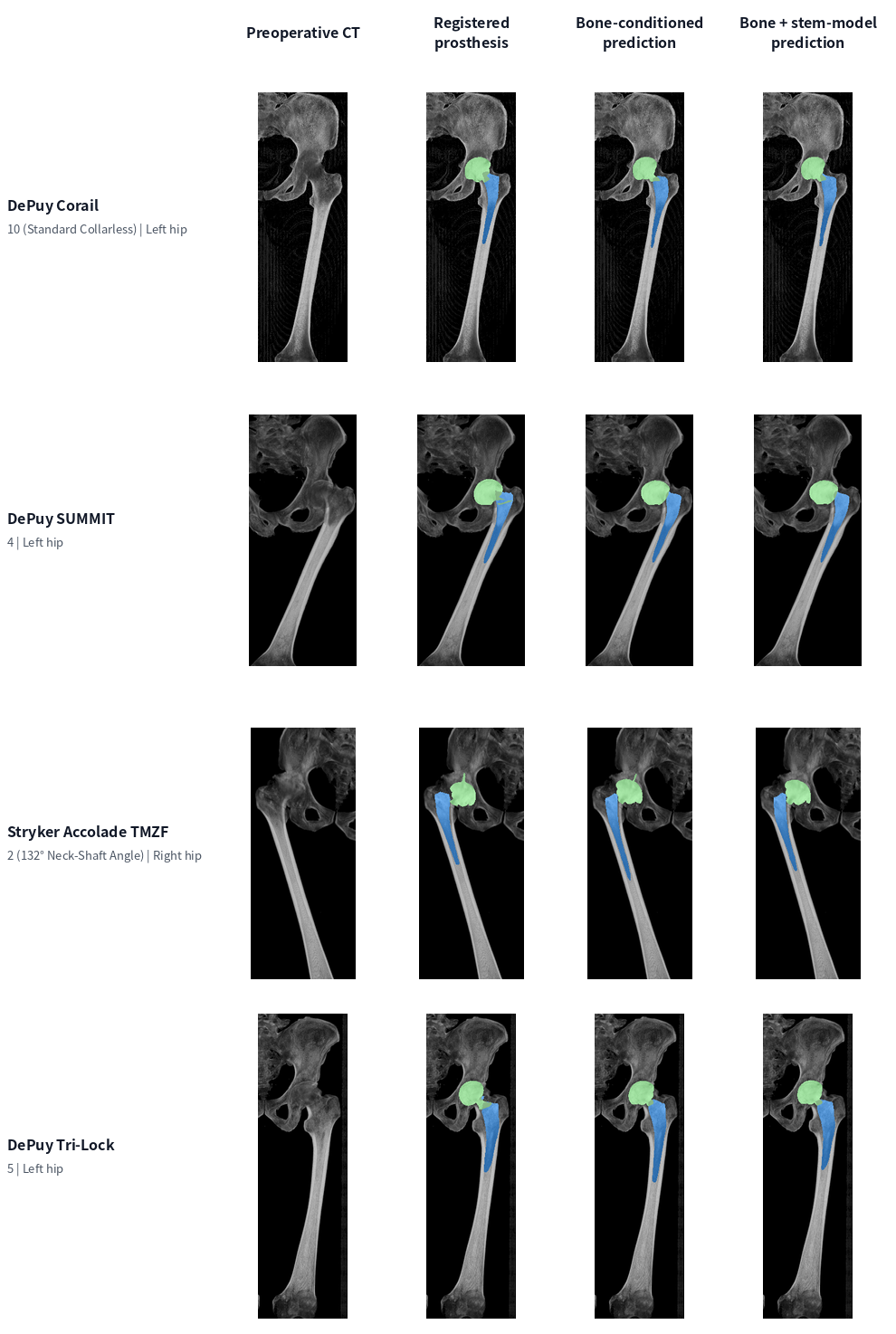}
\caption{\textbf{Patient-matched generation across major femoral stem designs.} Each row represents one recorded postoperative stem model: Corail, SUMMIT, Accolade TMZF, and Tri-Lock from top to bottom. Columns show the preoperative CT, the actual prosthesis registered to preoperative space, bone-only generation, and generation conditioned on bone and the recorded stem model. The two generated results use the same initial noise; neither receives stem size or other component parameters. Green and blue denote the acetabular and femoral prosthesis channels, respectively.}
\label{fig:cross-model}
\end{figure}

\begin{figure}[H]\ContinuedFloat
\centering
\includegraphics[width=0.92\textwidth]{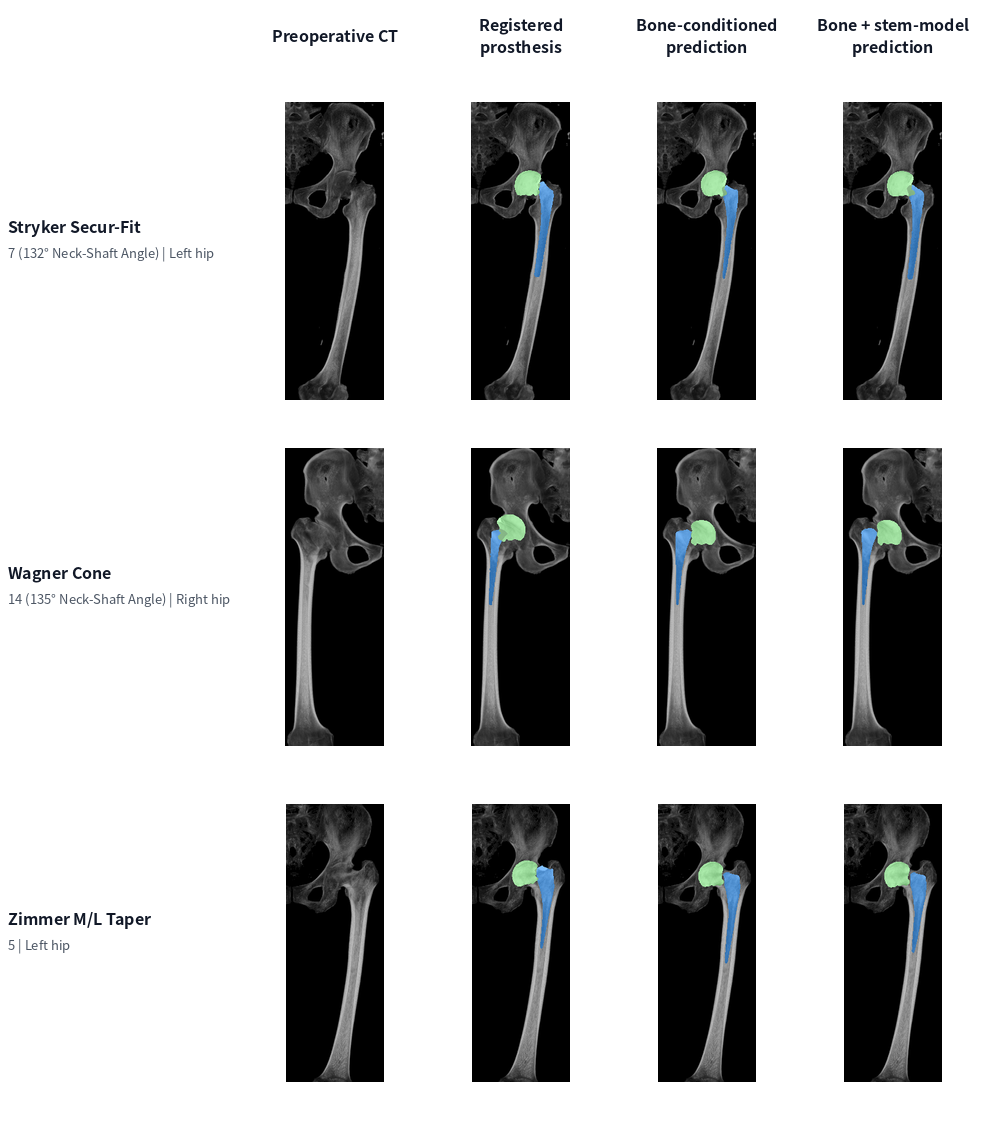}
\caption[]{\textbf{Patient-matched generation across major femoral stem designs (continued).} Secur-Fit, Wagner Cone, and M/L Taper are shown from top to bottom. Column definitions, conditioning modes, and colour coding are identical to those in Fig.~\ref{fig:cross-model}.}
\end{figure}

\subsection{Bone--Prosthesis Fit and Fill}
\begin{figure}[H]
\centering
\includegraphics[width=\textwidth]{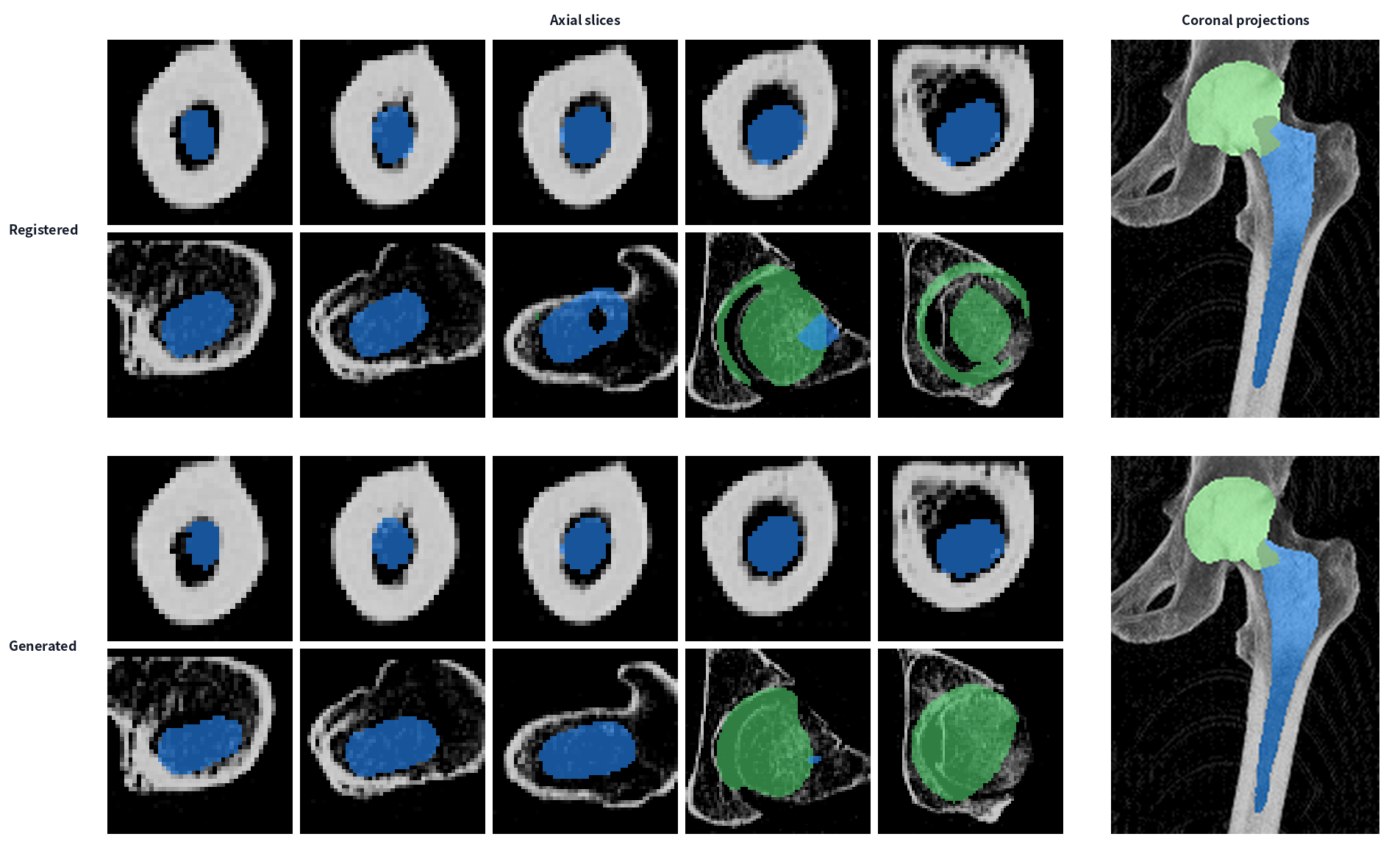}
\caption{\textbf{Bone--prosthesis fit and fill on consecutive sections and coronal projections.} The upper half shows the actual prosthesis registered to preoperative space; the lower half shows the generated prosthesis conditioned on bone and the Corail stem model. Axial sections on the left progress from the distal stem to the acetabulum, while coronal projections on the right show overall component position and alignment. The femoral channel in blue demonstrates fit within the effective canal, distal stem geometry, and respect for the calcar boundary; the acetabular channel in green demonstrates the spatial relationship between the cup and the acetabular wall.}
\label{fig:axial-series}
\end{figure}

Fit and fill at the bone--prosthesis interface are central to evaluating prosthesis geometry and placement. The effective canal is the anterolateral region of the proximal medullary canal that accommodates the stem and contributes to fixation, with its posteromedial boundary defined by the femoral calcar\cite{wang1994effective}. Consecutive axial sections from the same patient show that the generated stem did not simply occupy the entire canal. Proximally, it entered the effective canal and expanded along the regions intended for contact without encroaching beyond the calcar boundary; distally, it retained a continuous, regular taper rather than deforming unnecessarily to every local variation in the canal wall (Fig.~\ref{fig:axial-series}). Consecutive acetabular sections showed a complete cup whose outer surface closely followed the curvature of the acetabular wall. Greater fill is not inherently better, and the expected contact regions vary with stem design and fixation philosophy. A single overlap score or surface distance cannot fully characterise interface quality; review of consecutive sections and anteroposterior projections by clinical experts therefore remains important. The observed selective proximal fit and fill, respect for the calcar boundary, and regular distal geometry are consistent with the requirements for initial stability and avoidance of cortical impingement in cementless stems.

The coronal anteroposterior projections further show the medial border of the generated stem closely following the dense principal compressive trabecular system that extends inferiorly from the medial femoral neck, while the distal stem retains a regular profile. Piecewise CT intensity normalisation increased contrast across the transition from cancellous bone to cortex, allowing the canal boundary, cortex, and load-bearing trabeculae to act as distinct spatial cues. Truncated distance-field supervision concentrated learning near the prosthesis surface and reduced the influence of remote background voxels. Together, these choices encouraged region-specific fit and fill rather than prediction of only a coarse component centre, axis, or canal occupancy.

\subsection{Conditional Distribution Consistency}
Retrospective postoperative CT records one plan implemented from among several potentially acceptable solutions, rather than a unique ground truth for the same preoperative anatomy. Voxel overlap or surface distance to that single clinical solution would count other plausible candidates as errors and is therefore insufficient for evaluating a probabilistic generator. A more appropriate criterion is conditional distribution consistency: repeated samples for one anatomy should remain concentrated in component position, alignment, principal bone--prosthesis interfaces, and overall scale, while retaining limited variation in local contour and length; across patients, the generated distribution should remain consistent with the clinical distribution of actual postoperative prostheses.

\begin{figure}[H]
\centering
\includegraphics[width=0.95\textwidth]{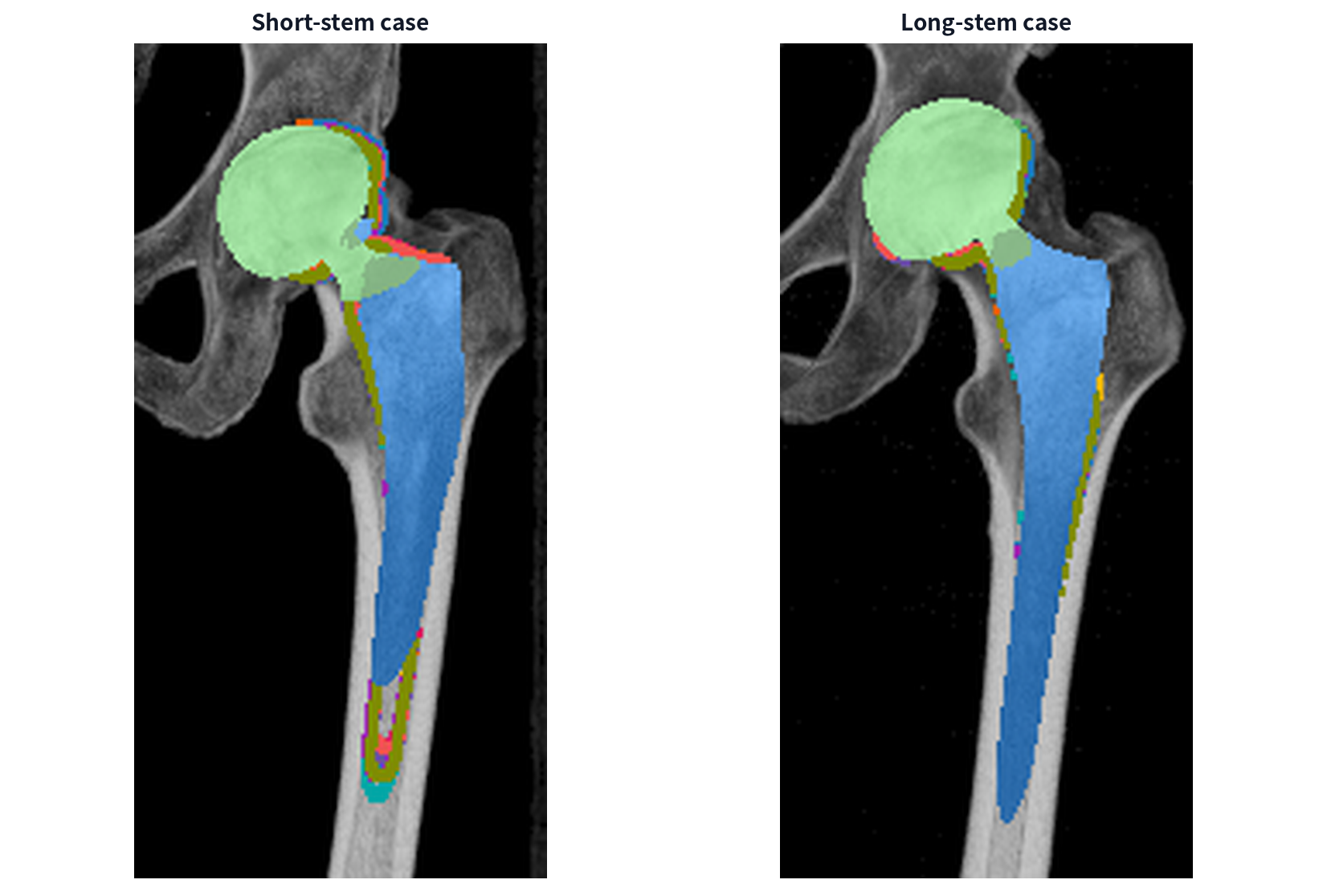}
\caption{\textbf{Within-patient bone-conditioned distribution relative to the clinical solution.} A short-stem case is shown on the left and a long-stem case on the right. Green and blue solids represent the actual acetabular and femoral components registered to preoperative space. Coloured lines delineate portions of generated projections extending beyond the actual prosthesis under different initial noise samples; overlapping regions are not redrawn. Inference uses the preoperative bone condition only, without prosthesis parameters.}
\label{fig:diversity}
\end{figure}

Repeated sampling across validation cases showed that most generated regions overlapped the registered actual prosthesis. Non-overlapping contours were concentrated near the cup rim, proximal stem boundary, and stem tip, forming a narrow distribution band (Fig.~\ref{fig:diversity}). Across cases, cup and stem position and stem alignment remained stable, whereas local shape differences represented neighbouring solutions permitted by the same anatomy. Diversity and convergence are therefore complementary: random initial noise explores local variation within the conditional distribution, while spatial bone conditioning anchors the common anatomical position and principal geometric extent. For a generative planning model, this distribution-level consistency is more informative than requiring every sample to reproduce a single postoperative solution.

\section{Discussion}
This study reformulates THA planning from discriminative prediction of a single model or placement into conditional generation of a three-dimensional prosthesis distribution. THA-Flow produced complete acetabular and femoral geometry across cases representing seven major stem models. Component position and alignment adapted to individual anatomy, the principal bone--prosthesis interfaces remained selectively matched, and repeated samples exhibited limited local variation under the same bone constraint. These findings indicate that the model learned a patient-matched conditional geometry distribution rather than a fixed template.

This formulation differs from two established pathways. Systems such as AIHIP use deep learning for segmentation and landmark detection before applying measurements, rules, and a standard prosthesis library to derive a single plan\cite{chen2022aihip}. THA-Net introduced generative modelling to THA, but generates simulated two-dimensional postoperative images\cite{rouzrokh2024thanet}. THA-Flow instead confines generation to an implicit three-dimensional field around the prosthesis surface, without redrawing bone, soft tissue, or metal artefacts. The output can be overlaid directly on the original preoperative CT or recovered as a solid from its zero level set, making it a more direct geometric object for surgical planning. To our knowledge, this is the first application of generative AI to three-dimensional planning for THA.

Reliable supervision in preoperative space is fundamental to the task. Because the pelvis and femur change pose independently between examinations, separate registration prevents joint motion from being confounded with variation in prosthesis placement. The dual-channel representation returns the acetabular components with the pelvis and the stem with the femur. It separates patient-matched generation from joint reduction at the neck while preserving the postoperative hip centre and head--cup spatial relationship as references for subsequent reduction and assembly.

Strong spatial conditioning and truncated distance-field supervision jointly shape the local bone--prosthesis relationship. Introducing the bone latent from the first convolution allows the canal wall, femoral calcar, load-bearing trabeculae, and acetabular wall to constrain geometry directly. Generated stems consequently showed selective fit and fill within the effective canal rather than indiscriminate canal occupation. Because appropriate contact regions vary with stem design and fixation philosophy, neither overlap nor surface distance alone fully captures interface quality. Moreover, postoperative CT records only one of several viable clinical solutions. Repeated sampling preserved component position, alignment, and the principal interfaces while allowing limited boundary variation, supporting conditional distribution consistency rather than agreement with a single postoperative result as the more appropriate concept of generative accuracy.

Structured parameters provide an additional level of control. Specifying a stem model modifies stem length, proximal width, and lateral profile while preserving anatomical fit, indicating that discrete product information can serve as an additional control over the conditional geometry distribution.

Several known limitations remain. First, the single-centre cohort may limit generalisability across patient populations, imaging protocols, and prosthesis inventories. Second, the strong spatial condition introduced by channel-wise bone concatenation deliberately prioritises patient matching, but consequently weakens the ability of model and size conditions to preserve standard product geometry. When a specified model and size are poorly compatible with the patient's anatomy, the generated result may represent an anatomically adapted deformation of that design rather than a faithful reproduction of its catalogue geometry. Finally, the model outputs continuous geometry rather than a directly actionable commercial model and size. Bone-only results therefore require geometric measurement or retrieval against a standard prosthesis library to identify the nearest available model and size, followed by component reduction and assembly to complete the surgical plan.

\section{Methods}
\subsection{Two-Stage Training Architecture}
\begin{figure}[H]
\centering
\includegraphics[width=\textwidth]{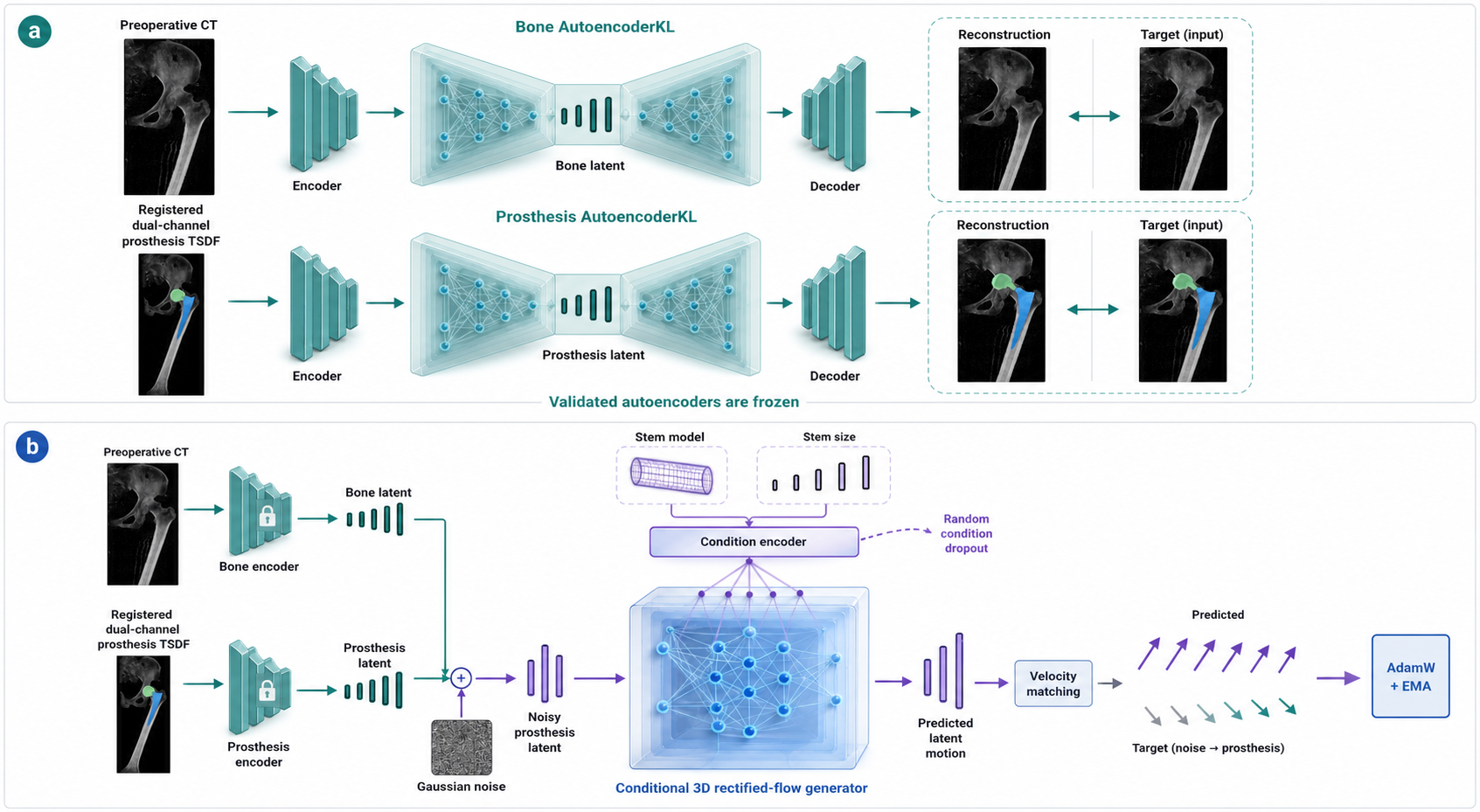}
\caption{\textbf{Latent pretraining and conditional rectified-flow matching.} \textbf{a}, Bone and prosthesis autoencoders (AutoencoderKL) learn latent representations of preoperative CT and dual-channel prosthesis \tsdf{}, respectively, and are frozen after reconstruction training. \textbf{b}, The bone latent is concatenated channel-wise with the current prosthesis path state. Stem model, size, and component parameters are encoded as semantic tokens. A three-dimensional UNet learns the conditional rectified flow from Gaussian noise to prosthesis geometry by velocity matching.}
\label{fig:training-architecture}
\end{figure}

Training used two stages: latent-space pretraining and conditional rectified-flow matching (Fig.~\ref{fig:training-architecture}). Separate AutoencoderKL models were first trained for bone and prosthesis geometry. The same prosthesis encoder--decoder processed the acetabular and femoral \tsdf{} channels independently, after which their two four-channel encodings were concatenated into the eight-channel prosthesis latent $y$. After reconstruction was validated, both autoencoders were frozen. Linear path states $y_\tau$ were sampled between $y$ and Gaussian noise $\epsilon$, concatenated voxel-wise with the bone latent $x$, and passed to a three-dimensional UNet trained to predict the prosthesis latent velocity $y-\epsilon$. Six structured prosthesis parameters were mapped by a separate encoder to condition tokens and injected through cross-attention at the deepest network level. Random condition dropout enabled one generator to learn unconditional, bone-conditioned, and partially or fully parameter-conditioned distributions.

\subsection{Study Design and Ethics}
This single-centre retrospective imaging study analysed CT examinations and operative records acquired during routine clinical care and involved no additional intervention. Eligible cases had temporally proximate preoperative and postoperative CT examinations from the same primary THA episode. Direct identifiers were removed before processing, and no research image contains an original accession number, patient name, or facial information.

\subsection{Cohort Construction and Data Split}
DICOM series were converted to three-dimensional NIfTI volumes, and axial and coronal digitally reconstructed projections were generated for rapid screening. Preoperative and postoperative examinations from the same operation were matched manually using examination dates, operative side, and imaging findings. Derived series, insufficient scan coverage, fractures within critical registration regions, and intentional intraoperative osseous alterations were excluded. TotalSegmentator was used only for coarse localisation of the operated hemipelvis and femur and identification of regions of interest, not to provide prosthesis supervision\cite{wasserthal2023totalsegmentator}.

Of 1,483 primary THA hips initially screened, 128 were excluded during CT pairing, prosthesis-record verification, rigid-registration quality control, or training-sample completeness checks. The final cohort comprised 1,355 hips from 1,149 patients, including 649 left and 706 right hips. The training, validation, and held-out test sets contained 1,188, 84, and 83 hips, respectively (Table~\ref{tab:split}). The test set remained sealed and was not used for model selection or figure preparation.

\begin{table}[H]
\centering
\caption{Distribution of included hips across the training, validation, and held-out test sets.}
\label{tab:split}
\begin{tabular}{lrrrr}
\toprule
Operated side & Training & Validation & Test & Total \\
\midrule
Left & 572 & 46 & 31 & 649 \\
Right & 616 & 38 & 52 & 706 \\
\midrule
Total & 1,188 & 84 & 83 & 1,355 \\
\bottomrule
\end{tabular}
\end{table}

Nine stem models used for model-stratified splitting covered 1,280 hips. The remaining 75 hips were not stratified: 73 lacked a reliable stem-model record and two belonged to extremely infrequent records. Validation and test cases were held out by stem-model strata to maintain coverage of major designs. Corail, SUMMIT, Accolade TMZF, and Tri-Lock together accounted for 1,106 hips, or 81.6\% of the cohort (Table~\ref{tab:models}).

\begin{table}[H]
\centering
\caption{Femoral stem model distribution in the included cohort.}
\label{tab:models}
\small
\begin{tabular}{lr@{\hspace{1.3cm}}lr}
\toprule
Stem model & Hips & Stem model & Hips \\
\midrule
DePuy Corail & 362 & Not model-stratified & 75 \\
DePuy SUMMIT & 343 & Wagner Cone & 53 \\
Stryker Accolade TMZF & 218 & Zimmer M/L Taper & 25 \\
DePuy Tri-Lock & 183 & Zimmer Wagner SL & 9 \\
Stryker Secur-Fit & 81 & DePuy S-ROM & 6 \\
\bottomrule
\end{tabular}
\end{table}

\subsection{Prosthesis Metadata and CT Verification}
Femoral stem model and size, acetabular cup, femoral head, and liner information were first extracted from itemised billing records for the same admission. Product catalogue numbers in billing entries were less susceptible to free-text transcription errors and were therefore used as the primary index. Publicly available manufacturer catalogues then mapped each catalogue number to stem model, model-specific size, cup outer diameter, head diameter, head offset, and liner offset. Fields that could not be established reliably from catalogue numbers and manufacturer documentation remained unlabelled and were excluded from parameter conditioning by field-specific masks during training.

Descriptive statistics covered nine stem models and 70 model-specific sizes. Cup outer diameter and head diameter were recorded or verified from CT for all 1,355 hips, with ranges of 40--62 mm and 22--40 mm, respectively. Cup diameters were concentrated at 48, 50, 52, and 54 mm (306, 293, 263, and 202 hips), while head diameters were predominantly 32 and 36 mm (611 and 572 hips). Head offset was labelled in 1,290 hips and ranged from $-5$ to $+8.5$ mm, with $+5$ and $+1.5$ mm accounting for 399 and 376 hips. Liner offset was labelled in 1,277 hips: 1,091 had 0 mm and 186 had $+4$ mm. Liner material was retained only as a descriptive variable; ceramic and polyethylene liners accounted for 777 and 578 hips and were not used as generative conditions. Missing fields naturally contributed partially conditioned examples during training.

\begin{figure}[H]
\centering
\includegraphics[width=0.98\textwidth]{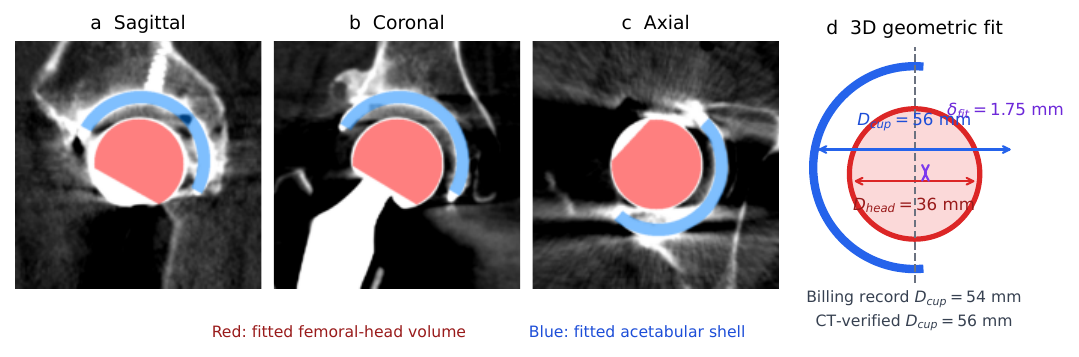}
\caption{\textbf{Verification of prosthesis component parameters on postoperative CT.} \textbf{a--c}, Candidate femoral-head spheres and hemispherical cup shells projected onto sagittal, coronal, and axial sections; fitting scores were computed over the complete three-dimensional volume. \textbf{d}, Definitions of cup outer diameter $D_{\mathrm{cup}}$, head diameter $D_{\mathrm{head}}$, and fitted centre offset $\delta_{\mathrm{fit}}$ along the cup axis. In this case, the billed cup diameter of 54 mm was corrected to a CT-measured diameter of 56 mm; $\delta_{\mathrm{fit}}$ was used only for cup-shell localisation.}
\label{fig:measurement}
\end{figure}

Cup outer diameter and head diameter were further verified three-dimensionally on postoperative CT. Using the billing record as an initial estimate, a 0.2-mm isotropic volume was resampled around the femoral head centre, and candidate head spheres and hemispherical cup shells were projected onto sagittal, coronal, and axial sections. NVIDIA Warp (the \texttt{warp-lang} Python package)\cite{macklin2022warp} kernels measured high-density metal occupancy inside and outside each candidate geometry throughout the full volume while jointly optimising head centre, cup axis, cup diameter, and fitting offset. An interactive Streamlit interface displayed all three orthogonal planes and allowed position, orientation, and dimensions to be reviewed in increments of 0.25 mm or $0.25^\circ$. In the example in Fig.~\ref{fig:measurement}, a billed cup diameter of 54 mm was corrected to the CT-supported value of 56 mm. Head and liner offsets retained their catalogue-derived values; the image-derived liner offset was used only for cup-shell localisation and not as a conditioning label.

\subsection{Independent Pelvic and Femoral Registration}
Metal artefacts from the postoperative cup and stem contaminate adjacent bone surfaces, precluding use of the complete postoperative bone surface. On the pelvic side, the inferior ischium provided an initial extent alignment, after which surfaces common to both scans, remote from cup artefacts, and unaffected by surgery were sampled. On the femoral side, volumes were aligned by their proximal extent; the lesser trochanter and adjacent shaft served as the principal rotation-specific anatomy, while the femoral neck osteotomy and metal-contaminated proximal regions were excluded. Both bones were registered using rigid iterative closest point optimisation without scaling or reflection\cite{besl1992icp}.

Separate transformation matrices were estimated for the acetabular and femoral sides to accommodate relative hip motion between examinations. Matrices estimated in local regions of interest were combined with crop offsets to recover transformations in the global physical coordinates of the original CT. Registration was not accepted on the basis of the ICP root-mean-square distance alone. Anteroposterior and lateral overlays were generated automatically for manual review of key bone surfaces, sampled points, and metal contours. Cases with fractures, subtrochanteric osteotomy, or other osseous changes between the lesser trochanter and stem tip were excluded. Additional fixation hardware was permitted when it did not compromise structural correspondence in this critical region.

\begin{figure}[!htbp]
\centering
\includegraphics[width=\textwidth]{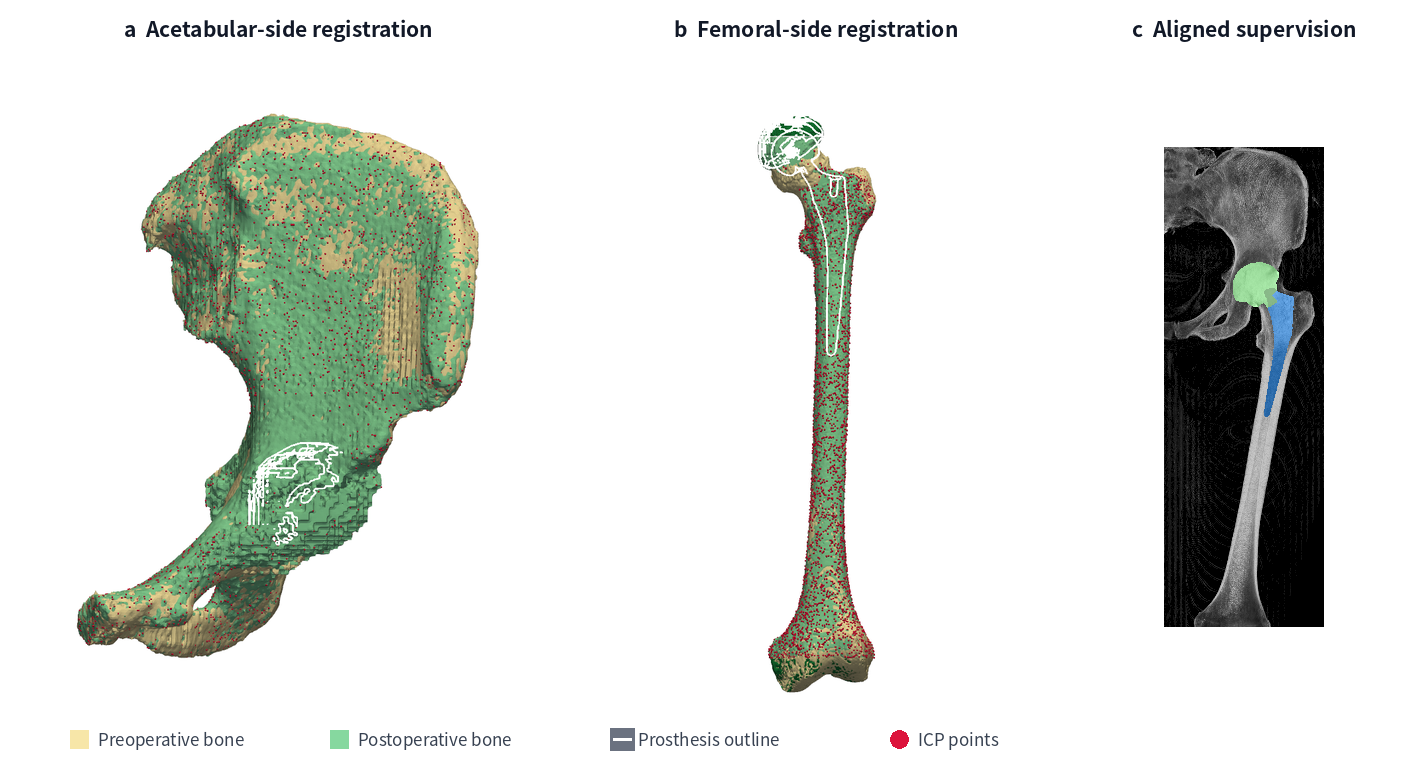}
\caption{\textbf{Independent rigid registration of the pelvis and femur establishes supervision in preoperative space.} \textbf{a,b}, Preoperative and postoperative bone surfaces and ICP samples for the pelvic and femoral registrations. Pale yellow and pale green indicate preoperative and postoperative isosurfaces, red points indicate postoperative surface samples selected away from metal artefacts and surgically altered regions, and white lines indicate postoperative prosthesis contours. \textbf{c}, The acetabular and femoral components are returned independently to preoperative CT space using their respective rigid transformations.}
\label{fig:registration}
\end{figure}

\subsection{Watertight Reconstruction and Dual-Channel \tsdf{}}
Metal was identified on postoperative CT using a high threshold of approximately 2,700 HU. Morphological closing and hole filling reduced false internal cavities caused by photon-starvation darkening within the femoral head. CUDA-accelerated DiffDMC reconstructed a closed, watertight surface from the metal voxels, and this solid surface defined the zero level set of the \tsdf{}. Distances were truncated at $\pm5$ mm, with positive values inside and negative values outside. The acetabular channel contained the cup and femoral head, while the femoral channel contained the stem; the channels were separated at the stem neck.

\begin{figure}[!htbp]
\centering
\includegraphics[width=0.96\textwidth]{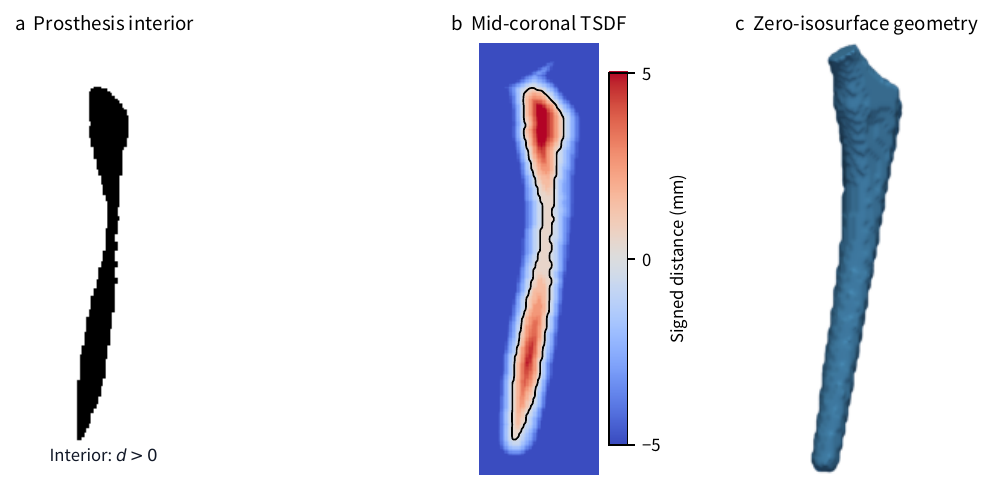}
\caption{\textbf{\tsdf{} representation of prosthesis geometry.} \textbf{a}, Solid femoral stem extracted from postoperative CT on a mid-coronal section. \textbf{b}, Continuous signed distance field on the same section; the black line marks the zero level set and distances are truncated at $\pm5$ mm. \textbf{c}, Closed, watertight stem surface recovered from the three-dimensional zero level set.}
\label{fig:tsdf}
\end{figure}

Unlike a binary mask, a \tsdf{} provides a continuous distance gradient on both sides of the surface while limiting the contribution of remote background voxels to the loss. A mid-coronal section displays the full stem length, and the three-dimensional solid can be recovered directly from the zero level set (Fig.~\ref{fig:tsdf}).

All images were resampled to 1-mm isotropic voxels, and each spatial dimension was padded to a multiple of 32. Preoperative CT underwent piecewise linear density normalisation. The transition from cancellous bone to cortex, spanning 150--650 HU, was mapped to the full intensity interval from $-1$ to 0; 650--1,150 HU and 1,150--3,150 HU were mapped to 0--0.5 and 0.5--1, respectively. This mapping enhanced density transitions among the canal, trabecular bone, and cortex. Training volumes covered the intersection of the preoperative scan and registered postoperative extent, with additional margins for the periprosthetic \tsdf{}. Spatial dimensions ranged from $128\times128\times192$ to $256\times352\times640$, with axis-wise medians of $160\times160\times512$. Voxel-wise operations including nearest-neighbour search, artefact-aware sampling, resampling, and distance-field computation were implemented as just-in-time compiled NVIDIA Warp GPU kernels. This avoided Python-level iteration and repeated CPU--GPU transfers over large volumes and supported a unified, automated GPU pipeline for parameter verification, registration, resampling, and distance-field construction.

\subsection{Latent Autoencoders and Reconstruction}
Separate KL-regularised autoencoders were trained for preoperative bone and prosthesis \tsdf{} volumes\cite{kingma2014vae}. Both used channel levels of 32, 64, and 128, two residual blocks per level, four-fold spatial downsampling, and four latent channels. Training used $128^3$ patches, whereas full volumes were encoded and decoded with Gaussian-weighted sliding windows. The encoder and decoder were fully convolutional to avoid inconsistency between patch boundaries and global self-attention.

Reconstruction objectives comprised L1 loss, weighted mean-squared error, and KL regularisation. The bone branch additionally used a PatchGAN least-squares adversarial objective and a three-dimensional MedicalNet perceptual loss\cite{chen2019med3d}. The prosthesis branch used Eikonal-gradient and solid-interior constraints. Models were trained with Adam; generator and discriminator learning rates were $10^{-4}$ and $2\times10^{-4}$, respectively, with $\beta=(0.5,0.9)$. Adversarial loss was disabled for the first five epochs. Each latent space was calibrated using its training-set global mean and scaling factor to approach zero mean and unit scale.

The bone and prosthesis autoencoders were selected at epochs 188 and 97, respectively. On the validation set, the bone branch achieved an L1 error of 0.007946, peak signal-to-noise ratio (PSNR) of 35.85 dB, and structural similarity index (SSIM) of 0.9802. Corresponding values for the prosthesis branch were 0.003313, 47.11 dB, and 0.9964 (Table~\ref{tab:vae}). The prosthesis branch achieved an interpolation Eikonal error of 0.013177 and a reconstruction Eikonal error of 0.007241, supporting the decoding of intermediate latents into continuous distance fields.

\begin{table}[H]
\centering
\caption{Best validation reconstruction performance of the bone and prosthesis autoencoders.}
\label{tab:vae}
\small
\begin{tabular}{lrrrr}
\toprule
Branch & Epoch & L1 & PSNR (dB) & SSIM \\
\midrule
Preoperative bone & 188 & 0.007946 & 35.85 & 0.9802 \\
Prosthesis \tsdf{} & 97 & 0.003313 & 47.11 & 0.9964 \\
\bottomrule
\end{tabular}
\end{table}

\begin{figure}[H]
\centering
\includegraphics[width=0.92\textwidth]{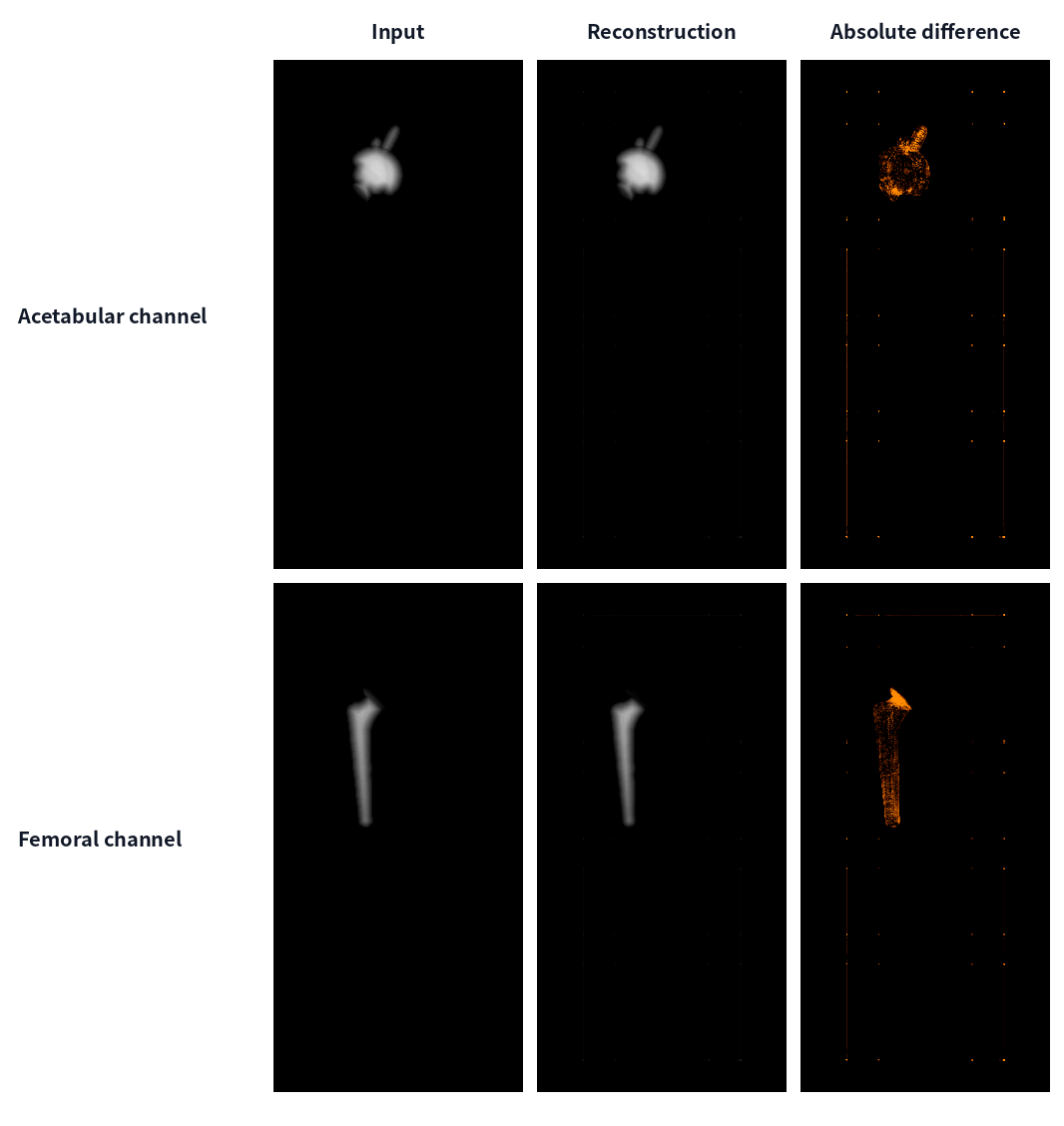}
\caption{\textbf{Latent reconstruction of the dual-channel prosthesis \tsdf{}.} The upper and lower rows show the acetabular and femoral channels, respectively. Columns show the original \tsdf{}, autoencoder reconstruction, and contrast-enhanced absolute error.}
\label{fig:vae-recon}
\end{figure}

\subsection{Bone-Conditioned Rectified Flow Matching}
Let $y$ denote the eight-channel prosthesis \tsdf{} latent, $x$ the four-channel preoperative bone latent, and $\epsilon\sim\mathcal{N}(0,I)$. Using normalised generative time $\tau$ from noise to data, the linear path is
\begin{equation}
y_{\tau}=(1-\tau)\epsilon+\tau y,\qquad \tau\sim\mathcal{U}(0,1),
\end{equation}
and the model regresses the velocity $y-\epsilon$:
\begin{equation}
\mathcal{L}_{\mathrm{FM}}=
\mathbb{E}_{y,x,\tau,\epsilon}
\left\|v_{\theta}([y_{\tau}\Vert x],\tau,c)-(y-\epsilon)\right\|_2^2,
\end{equation}
where $[\cdot\Vert\cdot]$ denotes channel-wise concatenation and $c$ is the optional structured prosthesis condition. This objective corresponds to single-stage rectified-flow matching\cite{liu2023rectifiedflow}.

The three-dimensional UNet received 12 channels and predicted eight, with channel levels of 96, 192, and 384 and two residual blocks per level. Self-attention was used only at the deepest level. The bone latent retained the same spatial dimensions as the prosthesis path state and entered as a channel condition from the first convolution. Structured parameters were represented by six 256-dimensional tokens and injected through cross-attention at the deepest level. The parameters comprised stem model, model-specific size, cup outer diameter, head diameter, head offset, and liner offset.

\subsection{Conditioning, Optimisation, and Inference}
Classifier-free condition dropout\cite{ho2022cfg} sampled six modes with equal probability: unconditional, all prosthesis parameters only, bone only, bone plus stem model, bone plus stem model and size, and bone plus all parameters. No classifier-free guidance amplification was applied during training. The generator was trained with AdamW at a learning rate of $10^{-4}$ using automatic mixed precision and a physical batch size of one. Gradients were accumulated dynamically according to voxel count to achieve an effective batch size of approximately 48. Exponential moving averages (EMA; decay 0.999) were maintained for both the generator and parameter encoder. At inference, Gaussian noise matching the target latent shape was integrated to the data endpoint with a small number of Euler steps and decoded by the prosthesis autoencoder using sliding windows.

The generator was trained for 1,000 epochs. Mean training velocity-field MSE decreased from 1.1524 to 0.0187, and the median validation MSE for bone-only conditioning over the final 100 epochs was 0.02099. EMA weights from epoch 999 were used for final inference. The large effective batch mitigated gradient variance arising from a physical batch size of one and variable volume dimensions, while EMA reduced sampling perturbations caused by individual parameter updates during prolonged optimisation.

\subsection{Computing Environment}
Data processing, model training, inference, and visualisation were performed in a Linux container. The compute node contained two NVIDIA RTX PRO 6000 Blackwell Max-Q Workstation Edition GPUs (97,887 MiB each), two Intel Xeon Platinum 8476C processors (96 physical and 192 logical cores in total), and 251 GiB of system memory. Each reported model-training or inference run used one GPU. The software environment comprised Python 3.13.13, PyTorch 2.11.0, CUDA 13.0, MONAI 1.5.2, ITK 5.4.6, and NVIDIA Warp 1.12.1 with GPU driver 580.173.02. NVIDIA Warp kernels supported registration sampling, distance computation, and resampling, while DiffDMC provided CUDA-accelerated watertight isosurface reconstruction.

\subsection{Evaluation of Generated Geometry}
All final results were generated with EMA weights. Primary visualisations were coronal anteroposterior projections and consecutive sections in preoperative CT space, with CT in greyscale and acetabular and femoral components in green and blue. Cross-design evaluation included validation cases with Corail, SUMMIT, Accolade TMZF, Tri-Lock, Secur-Fit, Wagner Cone, and M/L Taper stems. Each case underwent bone-only generation and generation conditioned jointly on bone and the recorded stem model. The two modes used identical initial noise and received neither stem size nor other component parameters, isolating the effect of stem-model conditioning.

Bone--prosthesis interface evaluation compared corresponding axial levels over the longitudinal extent shared by the actual and generated prostheses. Coronal projections were reviewed concurrently for stem alignment, fit and fill within the effective canal, respect for the calcar boundary, and the relationship between cup and acetabular wall. Conditional distribution evaluation repeated bone-only generation while varying the initial noise and used the registered actual prosthesis as a clinical reference. Portions of the generated projection extending beyond the actual projection were overlaid as differently coloured contours, while overlapping regions were not redrawn, thereby depicting the within-patient spatial distribution. L1, PSNR, SSIM, Eikonal error, and velocity-field MSE were used to verify latent reconstruction and training convergence, not as stand-alone measures of clinical planning quality.

\section*{Data Availability}
The raw and processed data cannot be made publicly available because of patient privacy, ethics requirements, and institutional data-governance restrictions.

\section*{Code Availability}
Source code is available at \url{https://github.com/doidio/nonahip}.

\section*{Author Contributions}
\begin{sloppypar}
Y.W. conceived the engineering architecture, developed the software and data-processing pipeline, trained and validated the models, analysed the results, prepared the visualisations, and drafted the manuscript. J.L. performed clinical case inclusion and exclusion, verified surgical and billing records, and reviewed the manuscript. J.S. performed data annotation, data curation, and quality control, and reviewed the manuscript. L.W. defined the clinical use scenarios, conducted expert review and clinical interpretation, supervised the clinical aspects of the study, and reviewed the manuscript. All authors approved the final manuscript.
\end{sloppypar}

\section*{Competing Interests}
Y.W. and J.S. are employees of Changzhou Jinse Medical Information Technology Co., Ltd. J.L. and L.W. declare no competing interests.


\begingroup
\small
\begin{thebibliography}{99}
\bibitem{colombi2019planning} Colombi, A., Schena, D. \& Castelli, C. C. Total hip arthroplasty planning. \textit{EFORT Open Rev.} \textbf{4}, 626--632 (2019). doi:10.1302/2058-5241.4.180075.
\bibitem{chen2022aihip} Chen, X. \textit{et al.} Development and validation of an artificial intelligence preoperative planning system for total hip arthroplasty. \textit{Front. Med.} \textbf{9}, 841202 (2022). doi:10.3389/fmed.2022.841202.
\bibitem{rouzrokh2024thanet} Rouzrokh, P. \textit{et al.} THA-Net: a deep learning solution for next-generation templating and patient-specific surgical execution. \textit{J. Arthroplasty} \textbf{39}, 727--733.e4 (2024). doi:10.1016/j.arth.2023.08.063.
\bibitem{rombach2022ldm} Rombach, R., Blattmann, A., Lorenz, D., Esser, P. \& Ommer, B. High-resolution image synthesis with latent diffusion models. In \textit{Proc. IEEE/CVF Conf. Computer Vision and Pattern Recognition} 10684--10695 (2022). doi:10.1109/CVPR52688.2022.01042.
\bibitem{wang2025meddiffusion} Wang, H. \textit{et al.} 3D MedDiffusion: a 3D medical latent diffusion model for controllable and high-quality medical image generation. \textit{IEEE Trans. Med. Imaging} \textbf{44}, 4960--4972 (2025). doi:10.1109/TMI.2025.3585372.
\bibitem{guo2025maisi} Guo, P. \textit{et al.} MAISI: medical AI for synthetic imaging. In \textit{Proc. IEEE/CVF Winter Conf. Applications of Computer Vision} 4430--4441 (2025). doi:10.1109/WACV61041.2025.00435.
\bibitem{zhao2026maisiv2} Zhao, C. \textit{et al.} MAISI-v2: accelerated 3D high-resolution medical image synthesis with rectified flow and region-specific contrastive loss. \textit{Proc. AAAI Conf. Artif. Intell.} \textbf{40}, 13088--13098 (2026). doi:10.1609/aaai.v40i15.38309.
\bibitem{wasserthal2023totalsegmentator} Wasserthal, J. \textit{et al.} TotalSegmentator: robust segmentation of 104 anatomic structures in CT images. \textit{Radiol. Artif. Intell.} \textbf{5}, e230024 (2023). doi:10.1148/ryai.230024.
\bibitem{besl1992icp} Besl, P. J. \& McKay, N. D. A method for registration of 3-D shapes. \textit{IEEE Trans. Pattern Anal. Mach. Intell.} \textbf{14}, 239--256 (1992). doi:10.1109/34.121791.
\bibitem{wang1994effective} Wang, Z. \& Dai, K. Geometric morphology of the femoral calcar and effective cavity of the proximal femur. \textit{Chin. J. Orthop.} \textbf{14}, 436--440 (1994) (in Chinese).
\bibitem{kingma2014vae} Kingma, D. P. \& Welling, M. Auto-encoding variational Bayes. In \textit{Proc. International Conference on Learning Representations} (2014).
\bibitem{chen2019med3d} Chen, S., Ma, K. \& Zheng, Y. Med3D: transfer learning for 3D medical image analysis. Preprint at \url{https://arxiv.org/abs/1904.00625} (2019).
\bibitem{liu2023rectifiedflow} Liu, X., Gong, C. \& Liu, Q. Flow straight and fast: learning to generate and transfer data with rectified flow. In \textit{Proc. International Conference on Learning Representations} (2023).
\bibitem{ho2022cfg} Ho, J. \& Salimans, T. Classifier-free diffusion guidance. Preprint at \url{https://arxiv.org/abs/2207.12598} (2022).
\bibitem{macklin2022warp} Macklin, M. Warp: a high-performance Python framework for GPU simulation and graphics. NVIDIA GPU Technology Conference (GTC) (2022). \url{https://github.com/NVIDIA/warp}.
\end{thebibliography}
\endgroup
\end{document}